\documentclass[10pt,twocolumn,letterpaper]{article}

\usepackage{iccv}
\usepackage{times}
\usepackage{epsfig}
\usepackage{graphicx}
\graphicspath{{img/}}  
\usepackage{amsmath}
\usepackage{amssymb}

\usepackage{enumitem}
\usepackage{makecell}
\usepackage{caption}

\usepackage[breaklinks=true,bookmarks=false]{hyperref}

\iccvfinalcopy 

\def\iccvPaperID{****} 

\begin{document}

\title{Satellite Imagery Feature Detection using \\ Deep Convolutional Neural Network: A Kaggle Competition}

\author{Vladimir Iglovikov\\
True Accord\\
{\tt\small iglovikov@gmail.com}
\and
Sergey Mushinskiy\\
Open Data Science\\
{\tt\small cepera.ang@gmail.com}
\and
Vladimir Osin\\
AeroState\\
{\tt\small resolutn@gmail.com}
}

\maketitle

\begin{abstract}
This paper describes our approach to the DSTL Satellite Imagery Feature Detection challenge \cite{dstl_competition} run by Kaggle. The primary goal of this challenge is accurate semantic segmentation of different classes in satellite imagery. Our approach is based on an adaptation of fully convolutional neural network for multispectral data processing.  In addition, we defined several modifications to the training objective and overall training pipeline, e.g. boundary effect estimation, also we discuss usage of data augmentation strategies and reflectance indices. Our solution scored third place out of 419 entries. Its accuracy is comparable to the first two places, but unlike those solutions, it doesn't rely on complex ensembling techniques and thus can be easily scaled for deployment in production as a part of automatic feature labeling systems for satellite imagery analysis.
\end{abstract}

\section{Introduction}
The significant increase of satellite imagery has given a radically improved understanding of our planet. Object recognition in aerial imagery enjoys growing interest today, due to the recent advancements in computer vision and deep learning, along with important improvements in low-cost high-performance GPUs. The possibility of accurately distinguishing different types of objects in aerial images, such as buildings, roads, vegetation and other categories, could greatly help in many applications, such as creating and keeping up-to-date maps, improving urban planning, environment monitoring, and disaster relief. Besides the practical need for accurate aerial image interpretation systems, this domain also offers scientific challenges to the computer vision.

In this paper, we describe and analyze these challenges for the specific satellite imagery dataset from a Kaggle competition.  We explore the challenges faced due to the small size of the dataset, the specific character of data, and supervised and unsupervised machine learning algorithms that are suitable for this kind of problems. Our efforts  can be summarized as follows:
\begin{itemize}
\item We adapted fully convolutional network to multispectral input data and evaluated several data fusion strategies on semantic segmentation task of satellite images.

\item We introduced joint training objective that properly defines desired output for the segmentation task.

\item We analyze local and global boundary effect on overall performance of the segmentation pipeline.

\end{itemize}

\section{Related Work}
\label{section:related_work}
Semantic segmentation for images can be defined as the process of partitioning and classifying the image into meaningful parts, and classify each part at the pixel level into one of the pre-defined classes. The current success of deep learning techniques in computer vision tasks motivated researchers to explore such techniques for pixel-level classification tasks as semantic segmentation. The Convolutional Neural Networks (CNN) is the main supervised approached that successfully used for this task. The key advancement of these networks is the ability to learn appropriate feature representation in an end to end manner while avoiding creation of hand-crafted features which require too much tuning to make them work in a particular case.

The most successful state-of-the-art deep learning method is the Fully Convolutional Network (FCN) \cite{fcn}.  The main idea of this approach is to use a CNN as a powerful feature extractor while replacing the fully connected layers with convolution ones to output spatial maps instead of classification scores. Those maps are upsampled to produce dense per-pixel output. This method allows training CNN in the end to end manner for segmentation with input images of arbitrary sizes. This approach achieved a notable enhancement in segmentation accuracy over common methods on standard datasets like PASCAL VOC \cite{pascal}.

Our solution is based on modified fully convolutional neural network architecture called U-Net  \cite{unet}, that was previously used for the tasks of biomedical image segmentation. The U-Net architecture allows combining low-level feature maps with higher-level ones, which enables precise localization.  A large number of feature channels in upsampling part allows propagating context information to higher resolution layers.  This type of network architecture was specially designed to solve image segmentation problems effectively. Technical details of U-Net adaptation for discussed task provided in Section \ref{section:methods}.

\section{Methodology}
\label{section:methods}

\subsection{Data description}

During the competition, the dataset was split into public and private (hidden) parts. Public part was used for model evaluations during first stage of the competition while final models were evaluated on private test set.
The dataset consists of 57 images that were divided into train (25 images) and test (32 images) sets. Each image covers 1 square kilometer of the earth surface. Satellite images of the same area can be separated into several types: a high-resolution panchromatic, an 8-band image with a lower resolution (M-band), and a short-wave infrared (A-band) that has the lowest resolution of all. The detailed band description is provided in \autoref{section:sensor}.

\begin{figure}[!h]
	\captionsetup{justification=centering}
	\centering
	\includegraphics[scale=0.225]{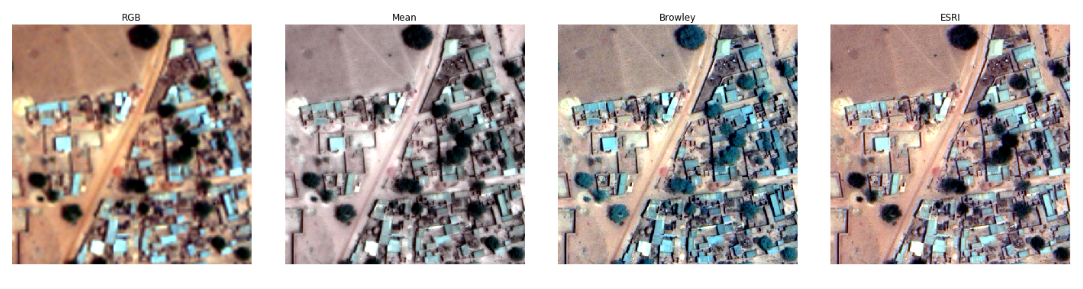}
	\caption{Panchromatic sharpening results for multispectral and panchromatic bands.}
	\label{fig:pan}
\end{figure}

RGB and M-band images partially overlap in the optical spectral range, because high-resolution RGB was itself reconstructed by panchromatic sharpening procedure. Panchromatic sharpening uses a higher-resolution panchromatic image to perform  fusion with a lower-resolution M-band image. The result produces a M-band image with the resolution of the panchromatic  where two rasters fully overlap. We apply several commonly used methods of panchromatic sharpening \cite{pansharp} to provided data (Fig.\ref{fig:pan}). Sharpened channels of M-band images can be used as alternative input to neural network.

The challenge is to assign one or more class labels to each pixel of the input image. The classes with additional description are presented in Table \ref{tab:classes}.

\begin{figure}[!h]
	\captionsetup{justification=centering}
	\centering
	\includegraphics[scale=0.4]{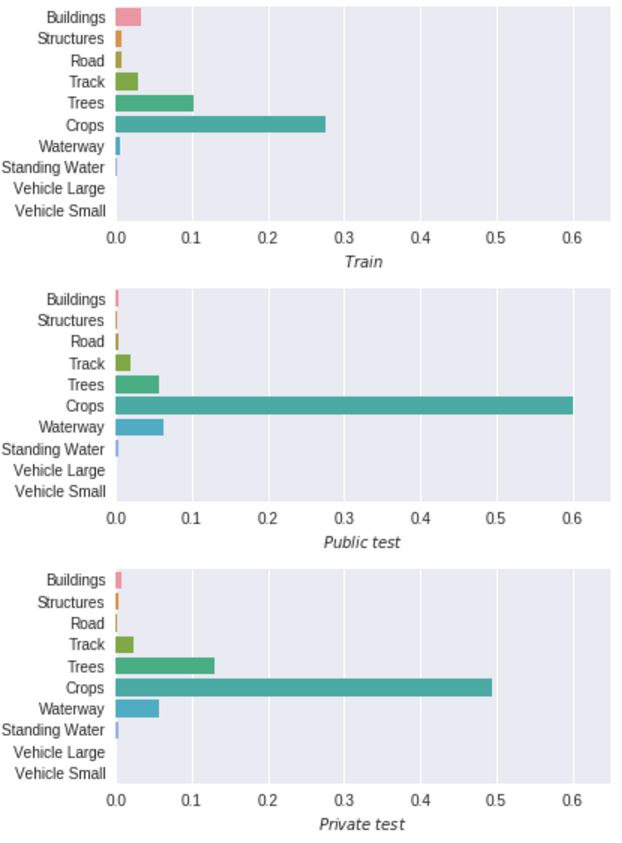}
	\caption{ Distributions of target classes for public and private parts of dataset.}
	\label{fig:cls_distr}
\end{figure}

The plots in Fig. \ref{fig:cls_distr} shows that target classes are heavily imbalanced within each set of images, e.g. one pixel of both large and small vehicle classes corresponds to a 60,000 pixels with crops. Therefore training a separate model per class provides much better results than single model for prediction of all classes. In addition, the distributions themselves vary significantly from one set of images to another. Also, the water classes were under-represented in the train set compared to the public and the private test sets. As a result, unsupervised methods described in Section \ref{subsec:reflectance} show better performance than neural network approaches.

\begin{table}[h!]
	\begin{center}
		\captionsetup{justification=centering}
		\begin{tabular}{|c|c|}
			\hline
			\textbf{Class}    & \textbf{Additional Description} \\ \hline
			Buildings      &  \makecell{large buildings, residential, \\ non-residential, fuel storage facilities, \\ fortified building} \\ \hline
			Structures     &  man-made structures	  \\ \hline
			Road           &  	 -  \\ \hline
			Track          & poor/dirt/cart tracks, footpaths/trails	  \\ \hline
			Trees 	       &   \makecell{woodland, hedgerows, groups of trees, \\ stand-alone trees}	  \\ \hline
			Crops 	       &  \makecell{contour ploughing/cropland, grain crops, \\ row (potatoes, turnips) crops}   \\ \hline
			Waterway       &   -   \\ \hline
			Standing water &   -   \\ \hline
			Vehicle Large  & \makecell{large vehicle (e.g. lorry, truck, bus), \\ logistics vehicle}      \\ \hline
			Vehicle Small  &  small vehicle (car, van), motorbike \\ \hline
		\end{tabular}
		\caption{Dataset contains  industrial and nature classes. \\ Described classes are varied in terms of shape and size.}\label{tab:classes}
	\end{center}
\end{table}

\subsection{Multispectral Sensor}
\label{section:sensor}

WorldView-3 satellite is one of the commercial satellites, which provide 31 cm panchromatic resolution, 1.24 m multispectral resolution, and 7.5 m resolution for short-wave infrared channels \cite{bands}. Multispectral images enable to extract important features that beyond human vision. For example, the near infrared wavelength is usually used to separate vegetation varieties and conditions due to strong reflection in this range of electromagnetic spectrum that vegetation provides. Besides, the color depth of such images is 11 and 14-bit instead of commonly used 8-bit. From a neural network perspective it is better, because each pixel carries more information, while it creates additional steps for proper visualization.

The multispectral bands can be used for recognition of specific classes of object:

\begin{itemize}
	\item \textbf{Coastal (400-452 nm)}. This band senses deep blues and violets.
	 It is also called the coastal/aerosol band, after its two main uses: imaging shallow water, and tracking fine particles like dust and smoke.

	\item \textbf{Blue (448-510 nm)}. This band senses normal blues. It provides increased penetration of water bodies and also capable of differentiating soil and rock surfaces from vegetation and for detecting cultural features.

	\item \textbf{Green (518-586 nm)}. This band senses greens. Because it covers the green reflectance peak from leaf surfaces, it has separated vegetation (forest, croplands with standing crops) from soil. In this band urban areas, roads and highways have appeared as brighter tone, but forest, vegetation, croplands with standing crops have appeared as dark (black) tone.

	\item \textbf{Yellow (590-630 nm)}. This band senses in a strong chlorophyll absorption region and strong reflectance region for most soils. It has separated vegetation and soil. But it could not separate water and forest. Forest land and water both have appeared in dark tone. This band has highlighted barren lands, urban areas, street pattern in the urban area and highways. It has also separated croplands with standing crops from bare croplands with stubble.

	\item \textbf{NIR (772-954 nm)}.This band measures the near infrared. This part of the spectrum is especially important for ecology purposes because healthy plants reflect it. Information from this band is important for major reflectance indexes, such as NDVI \cite{ndvi}, which allow to measure specific characteristics more precisely.

	\item \textbf{SWIR (1195-2365 nm)}. This band cover different slices of the shortwave infrared. They are particularly useful for telling wet earth from dry earth, and for geology: rocks and soils that look similar in other bands often have strong contrasts in this band.
\end{itemize}

\begin{table*}[th]
	\centering
	\begin{tabular}{|c|c|c|c|}
		\hline
		\textbf{Bands} & \textbf{Spectral Range} & \textbf{Resolution} & \textbf{Dynamic Range} \\ \hline
		Panchromatic & 450-800 nm & 0.31 m & 11 bits/pixel\\ \hline
		Multispectral 8-bands & 400-1040 nm & 1.24 m & 11 bits/pixel \\ \hline
		SWIR 8-bands & 1195-2365 nm & 7.5 m & 14 bits/pixel\\ \hline
	\end{tabular}
	\caption{Specifications for the WorldView-3.}
	\label{table:specs}
\end{table*}

\subsection{Reflectance indices}
\label{subsec:reflectance}

\begin{figure}[!h]
	\captionsetup{justification=centering}
	\centering
	\includegraphics[scale=0.27]{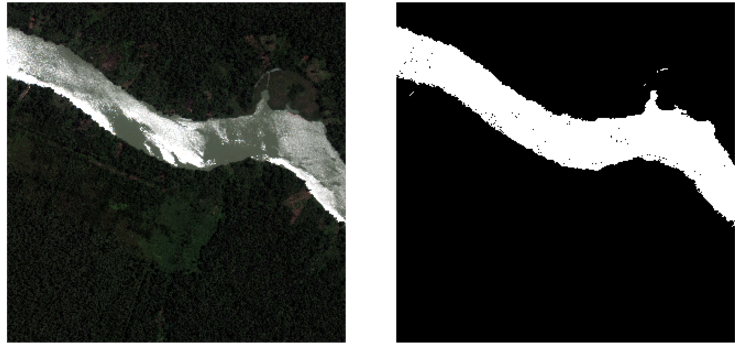}
	\includegraphics[scale=0.27]{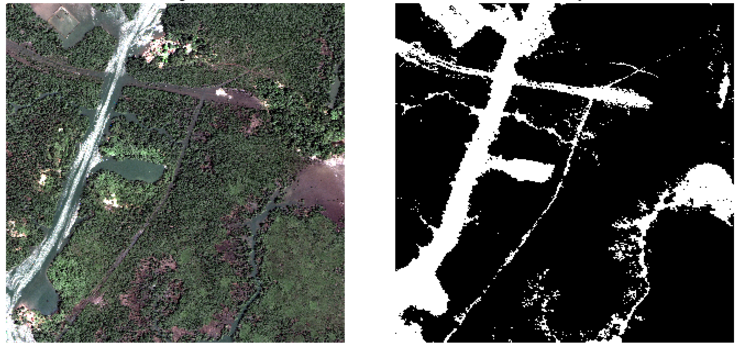}
	\caption{ Reflectance indexes (thresholded CCCI) showed best performance in segmentation of waterways.}
	\label{fig:ccci}
\end{figure}

Apparently, the fact that we have infrared and other channels from non-visible frequency range allows us to identify some classes purely from the pixel values, without any contextual information. Using this approach, the best results were obtained for water and vegetation classes. For instance, in our final solution both water classes were segmented using CCCI \cite{ccci} and NDWI reflectance indices, that defined as follows:

\begin{equation}
{\rm CCCI} = \frac{{\rm NIR} - {\rm RED}_{\rm edge}}{{\rm NIR} + {\rm RED}_{\rm edge}}\times{\frac{{\rm NIR} + {\rm RED}}{{\rm NIR} - {\rm RED}}}
\end{equation}

\begin{equation}
{\rm NDWI} = \frac{{\rm GREEN} - {\rm NIR}}{{\rm GREEN} + {\rm NIR}}
\end{equation}

\noindent where both indices are represented as ratios of the difference and sum of pixel values in the green, red edge, red and infrared channels. The results shows high intensity values for waterways, but it also shows false positives for some buildings due to the relative similarity of the specific heat of metal roofs and water. Water classes segmentation results are presented in Fig. \ref{fig:ccci}.

We expected a deep learning approach to perform on par or even better than index thresholding and, in vegetation prediction, neural networks did indeed outperform indices. However, we found that indices allow us to achieve better results for under-represented classes such as waterways and standing water. In the provided images ponds were smaller than rivers, so we additionally thresholded our predictions by area of water body to distinguish waterways from standing water.

\subsection{Multispectral U-NET}

In general, U-Net architecture consists of contracting and expansive paths. The contractive path follows the typical convolution neural network architecture.
We use a batch normalization \cite{batchnorm} for convergence acceleration during training. In addition, instead of the rectified linear unit we use exponential linear unit (ELU) \cite{elu} as primary activation function, which is beneficial for learning and it helps to learn representations that are more robust to noise. The number of feature channels is doubled at each down-sampling step. Expansive path consists of upsampling operation of the feature map followed by convolution with half number of feature channels, concatenation with the corresponding feature map from contracting path, also followed by batch normalization and ELU.

\begin{figure*}[th]
	\centering
	\includegraphics[scale=0.4]{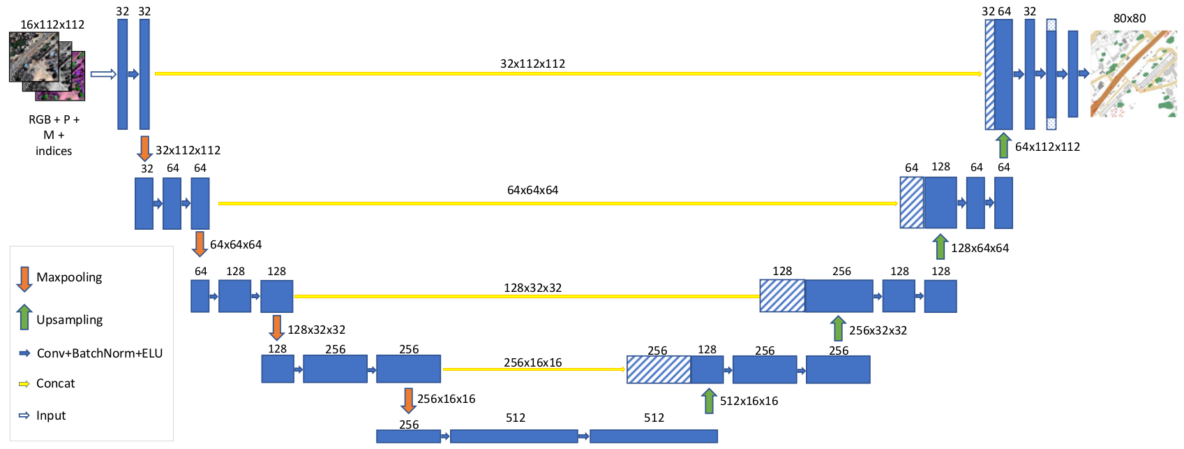}
	\captionsetup{justification=centering}
	\caption{Multispectral U-Net architecture consists of down-sampling and upsampling \\ parts with skip connections between them.  Combinations of multispectral and \\ panchromatic bands with main reflectance indices combined \\ into single tensor for input to the neural network. }
	\label{fig:unet}
\end{figure*}

As a primary input we perform early fusion of multispectral bands, reflectance indices and RGB channels, stacking them into single tensor. The full architecture is presented in Fig. \ref{fig:unet}.

Evaluation metric is the Jaccard index, also known as intersection over union, which can be interpreted as similarity  measure between a finite number of sets. Intersection over union for similarity measure between two sets $A$ and $B$ can be defined as following:

\begin{equation}
\begin{aligned}
J(A, B) = \frac{\mid A \cap B \mid }{\mid A \cup B \mid} &= \frac{\mid A \cap B \mid}{\mid A \mid + \mid B \mid - \mid A \cap B \mid} \\ \\
0 \le & J(A, B) \le 1
\end{aligned}
\end{equation}

The common loss function for classification tasks is categorical cross entropy, however in our case classes are not mutually exclusive and using binary cross entropy makes more sense. The binary cross entropy defined as follows:

\begin{equation} \label{eq:bintropy}
H = -\frac{1}{n}\sum_{i=1}^n{[y \log(\widehat{y})+(1-y)\log(1-\widehat{y})]}
\end{equation}

In order to get better results, training objective and evaluation metric should be as close as possible. However, the problem is that Jaccard Index is not differentiable. One can generalize it for probability prediction, which on the one hand, in the limit of the very confident predictions, turns into normal Jaccard and on the other hand is differentiable – allowing the usage of it in the algorithms that are optimized with gradient descent (\ref{eq:jacmod}).

\begin{equation} \label{eq:jacmod}
J_{m}(y, \widehat{y}) = \frac{1}{n} \sum_{i=1}^n \frac{y_i \cdot \widehat{y_i}}{y_i + \widehat{y_i} - y_i \cdot \widehat{y_i}}
\end{equation}

Thus, the joint loss function defined as combination of Eq.\ref{eq:bintropy} and Eq.\ref{eq:jacmod}:

\begin{equation}
L = H - \log J_{m}
\end{equation}

We used Nadam Optimizer (Adam with Nesterov momentum) \cite{nadam} and trained the network for 50 epochs with a learning rate of 1e-3 and additional 50 epochs with a learning rate of 1e-4. Each epoch was trained on 400 batches, each batch contained 128 image patches. Each batch was created randomly cropping $112 \times 112$ patches from original images. In addition each patch was modified by applying a random transformation from $Dih_4$ group \cite{D4}.

We also tried $224 \times 224$ patches but due to limited GPU memory this would significantly reduce the batch size from 128 to 32. Larger batches proved to be more important than a larger receptive field. We believe that it was due to the train set containing 25 images only, which differ from one another quite heavily. As a result, we decided to trade-off receptive field size in favor of a larger batch size.

\subsection{Boundary effects}

\begin{figure}[!h]
	\captionsetup{justification=centering}
	\centering
	\includegraphics[scale=0.2]{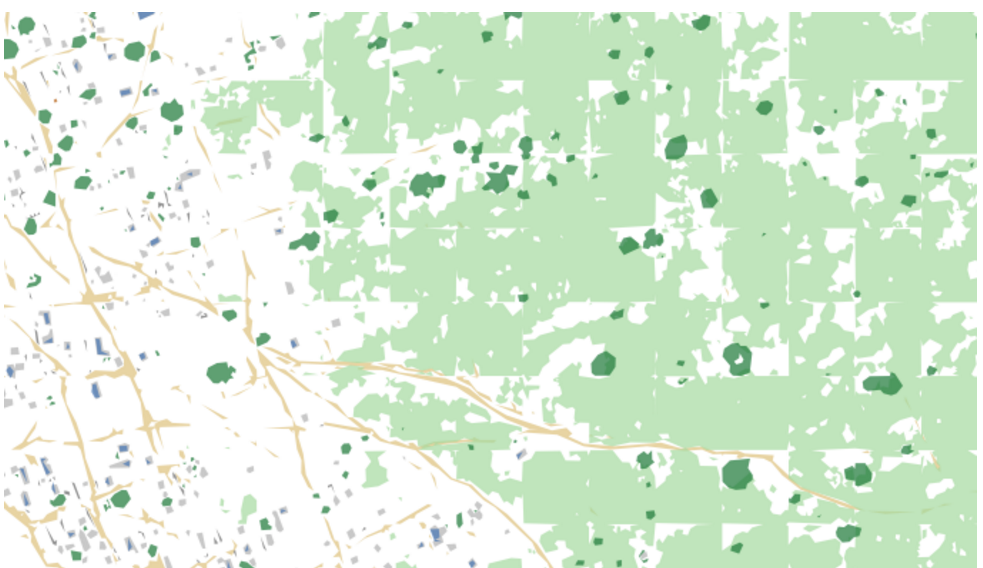}
	\caption{Image formed via patch-based predictions. This lead to boundary effects in a form of square structure near the edge of each patch.}
	\label{fig:boundarybad}
\end{figure}

During training procedure we prepared patches, cropping them from the original images, augment and feed into the neural network. However, the same patching strategy applied during prediction stage, lead to square structure in resulted image, as presented in Fig. \ref{fig:boundarybad}.

\begin{figure}[!h]
	\captionsetup{justification=centering}
	\centering
	\includegraphics[scale=0.25]{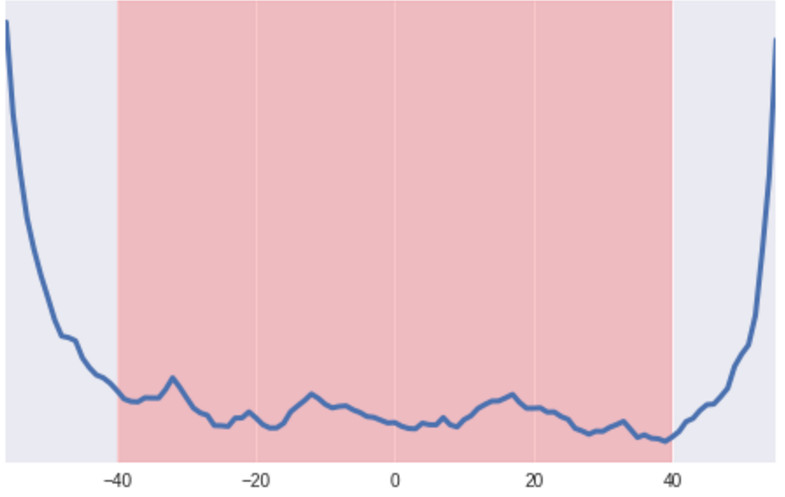}
	\caption{Prediction quality decreasing from the center of the patch to the edges, that showed as a logistic loss with respect to distance from center for building class.}
	\label{fig:boundary}
\end{figure}

The main reason of such square structure is that not all outputs in the Fully Connected Network are equally good. Number of ways to get from any input pixel to the central part of the output in a network is much higher than to get the edge ones. As a result, prediction quality is decreasing when you move away from the center. We tested this hypothesis for a few classes, e.g analysis for building class, presented in Fig. \ref{fig:boundary}.

One way to deal with such an issue was to make the predictions on overlapping patches, and crop them on the edges, but we came out with a better way. We added cropping layer to the output layers of our networks, which solved two main problems simultaneously:

\begin{enumerate}
	\item Losses on boundary artifacts were not back-propagated through the network;
	\item Edges of the predictions were cropped automatically.
\end{enumerate}

This trick slightly decreased computational time.

\begin{figure}[!h]
	\captionsetup{justification=centering}
	\centering
	\includegraphics[scale=0.21]{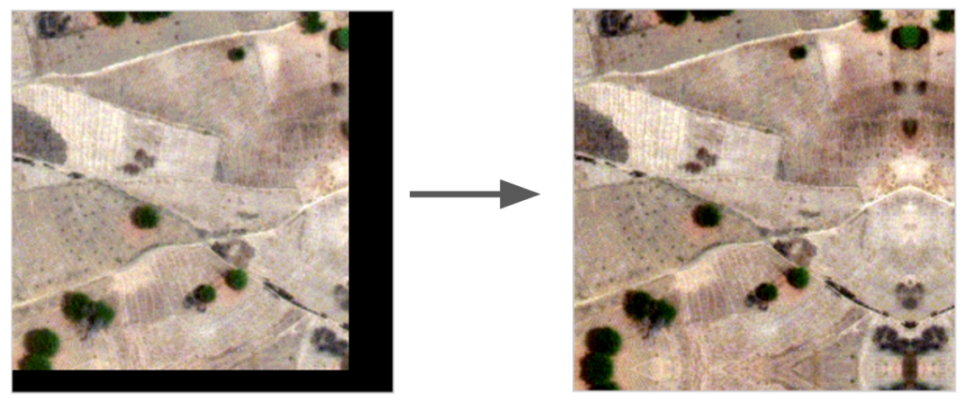}
	\caption{Reflections of the central part to the padded areas helps to prevent global boundary effect.}
	\label{fig:boundary2}
\end{figure}

To summarize, we trained a separate model for each of the first six classes. Each of them took a matrix with a shape $128 \times 16 \times 112 \times 112$ as an input and returned the mask for a central region of the input images $128 \times 1 \times 80 \times 80$ as an output. Besides, global boundary effects arised during partition of original $3600 \times 3600$ images into $112 \times 112$ tiles due to zero padding. This added some problems at the prediction time. For example, sharp change in pixel values from central to zero padded area was probably interpreted by a network as a wall of a building and as a result we got a layer of a building all over the perimeter in the predicted mask. We address this issue with the same trick as in the original U-net paper \cite{unet} (See Fig. \ref{fig:boundary2}).

\section{Results}

In conclusion, we would like to add that successful approach 
to above-mentioned problems allows to significantly improve the quality of final models. Our approach includes several steps, such as the adaptation of fully convolutional network to multispectral satellite images with joint training objective and analysis of boundary effects, reflectance indices. The final results summarized in Table \ref{tab:results}.

\begin{table}[h]
	\begin{center}

	\begin{tabular}{c|c|c}
\textbf{Class}    & \textbf{Public set} & \textbf{Private set} \\ \hline \hline
Buildings  & 0.7453 & 0.6290 \\
Structures & 0.1905 & 0.2015 \\
Road  & 0.8005 & 0.5605 \\
Track & 0.3281 & 0.3965 \\
Trees & 0.5018 & 0.6984 \\
Crops & 0.8251 & 0.8280 \\
Waterway & 0.9697 & 0.9131 \\
Standing water & 0.6081 & 0.5272 \\
Vehicle Large & 0.2964 & 0.0331 \\
Vehicle Small & 0.0186 & 0.00000 \\
		\end{tabular}
	\end{center}
	\captionsetup{justification=centering}
	\caption{Segmentation results for different classes in terms of intersection over union.}\label{tab:results}
\end{table}


\newpage

\begin{thebibliography}{20}

	\bibitem{fcn} Long, Jonathan, Evan Shelhamer, and Trevor Darrell. "Fully convolutional networks for semantic segmentation." Proceedings of the IEEE Conference on Computer Vision and Pattern Recognition. 2015.

	\bibitem{pascal} Everingham, Mark, et al. "The pascal visual object classes challenge: A retrospective." International Journal of Computer Vision 111.1 (2015): 98-136


	\bibitem{unet} Ronneberger, Olaf, Philipp Fischer, and Thomas Brox. "U-net: Convolutional networks for biomedical image segmentation." International Conference on Medical Image Computing and Computer-Assisted Intervention. Springer International Publishing, 2015.

	\bibitem{ndvi} Rouse Jr, et al. "Monitoring vegetation systems in the Great Plains with ERTS." (1974).

	\bibitem{ccci} Cammarano, Davide, et al. "Use of the canopy chlorophyl content index (CCCI) for remote estimation of wheat nitrogen content in rainfed environments." Agronomy journal 103.6 (2011): 1597-1603.

	\bibitem{nadam} Dozat, Timothy. Incorporating Nesterov momentum into Adam. Stanford University, Tech. Rep., 2015.[Online]. Available: http://cs229. stanford. edu/proj2015/054 report. pdf, 2015.

	\bibitem{bands} World View-3 Satellite Sensor Specifications.  \texttt{http://www.satimagingcorp.com/}

	\bibitem{batchnorm} Ioffe, Sergey, and Christian Szegedy. "Batch normalization: Accelerating deep network training by reducing internal covariate shift." arXiv preprint arXiv:1502.03167 (2015).

	\bibitem{elu} Clevert, Djork-Arné, Thomas Unterthiner, and Sepp Hochreiter. "Fast and accurate deep network learning by exponential linear units (elus)." arXiv preprint arXiv:1511.07289 (2015).

	\bibitem{pansharp} Padwick, Chris, et al. "WorldView-2 pan-sharpening." Proceedings of the ASPRS 2010 Annual Conference, San Diego, CA, USA. Vol. 2630. 2010.

	\bibitem{dstl_competition} \url{https://www.kaggle.com/c/dstl-satellite-imagery-feature-detection}

	\bibitem{D4} \url{https://en.wikipedia.org/wiki/Dihedral_group}

\end{thebibliography}
\end{document}